# Studying Positive Speech on Twitter

Marina Sokolova[1,4,+], Vera Sazonova[2,*], Kanyi Huang[1], Rudraneel Chakraboty[3,**], Stan Matwin[4]

[1]University of Ottawa, Ottawa, ON, Canada
[2]Kea Text, Montreal, QC, Canada
[3]Carleton University, Ottawa, ON, Canada
[4]Dalhousie University, Halifax, NS, Canada

**Abstract.** We present results of empirical studies on positive speech on Twitter. By positive speech we understand speech that works for the betterment of a given situation, in this case relations between different communities in a conflict-prone country. We worked with four Twitter data sets. Through semi-manual opinion mining, we found that positive speech accounted for < 1% of the data . In fully automated studies, we tested two approaches: unsupervised statistical analysis, and supervised text classification based on distributed word representation. We discuss benefits and challenges of those approaches and report empirical evidence obtained in the study.

## Introduction

Dynamic Social Impact Theory shows that overall influence a person experiences from others is a function of the strength, immediacy, and number of communications from other people [7]. Social network Twitter is the fastest growing medium of interpersonal communication. It is visible on public Internet and often cited by other media [24]. Persuasion and social influence that operate through the network affect millions of its users and the general public at large [6, 10, 20]. Social Mining and Social NLP studies became involved in analysis of sentiments and emotions exhibited on Twitter. Sentiments are studied in tweets related to Olympic Games 2010 in Vancouver [10], a major 2012 earthquake in Japan [20], the 2011 Spanish legislative elections, and the 2012 US presidential elections [1]. Positive, consolidation speech is a part of a multi-faceted peace-building process [28]. It contributes to the construction of a culture of peace to replace a structure of violence [15]. Although the promises of social networks in peace-building and post-conflict re-consolidations are palpable, considerable research is required to assess effect of consolidation speech in conflict-stressed environments [3].

Until now Text Data Mining community did not pay as much attention to positive speech as to offensive and hate speech. Whereas several types of offensive and hate speech were studied in Sentiment Analysis [17, 21, 26], we could not find Sentiment Analysis/Opinion Mining studies focused on consolidation aspects of speech or near-synonymous concepts. In this paper we report analysis of positive speech and present empirical results obtained on four Twitter data sets. The data has been collected in Kenya through UMATI project[1]. We worked with tweets written in English. Through semi-manual opinion mining, we found that positive speech accounted for < 1% of the data. This severe class imbalance was a major challenge for fully automated studies, as traditional data imbalance methods - under-sampling of majority class and over-sampling of minority class - do not work well with class imbalance of 1/99. Instead, we opted to test two fully automated methods: unsupervised topic allocation by Latent Dirichlet Allocation and supervised text classification based on distributed word representation. The approaches were applied independently. The obtained results of the two approaches collaborate each other. Further in the paper, we revise relevant work, identify the scope of the study, introduce the data sets, and report empirical results that support our approach; discussion concludes the paper.

---

**Relevant Work**

By speech we understand the means of communication used by humans to express ideas and thoughts by means of words and lexical constructions [27]. Positive speech works for the process of betterment of a given situation. It is context-dependent. On our case, data is collected from conflict-ridden zones in Kenya, where clashes happen along ethnical/religious lines. In this context, positive speech helps to build bridges between two different population groups or find common grounds between different opinions. Hence, consolidation is an important topic of positive speech. Documentary reports on reconstruction efforts belong to positive speech too. Expressions of sorrow and mourning for all victims represent another major type of positive speech in the given context as well as speech directly countering hate speech. The following tweet exempts are considered positive speech:
*we need to be united beyond the borders of religion*,
*[XX] has donated Sh3 million for the reconstruction of the Gikomba market*
*Dont celebrate over Makaburi's death! Be kind ,Life is interesting ,Life is unpredictable!*

The micro-bloging network Twitter has become a world-wide connector, having 284 million monthly active users and 500 million tweets sent per day[2]. Twitter data have become a fertile ground for many research projects: advanced search `Twitter, sentiment analysis, From 2011` yields > 5,500 hits on Google Scholar. In this section we review several papers that analyze negative and positive polarity of speech on Twitter.

Twitter users are keen to post negative sentiments and emotions more than they post positive sentiments and emotions [10]. In part, this imbalance can be explained by correlation between social anxiety and negative cognitive bias: people with negative cognitive bias tend to be more socially anxious, and vis-a-verse [13]. The negative balance can be exaggerated by the fact that Twitter users who tweet about politics tend to have extreme ideological preferences [1]. Offensive speech, i.e., speech containing offensive terms, is a special type of negative speech. Offensiveness of tweets can be estimated with a reasonable error rate of 12.4% obtained by SVM in binary offensive – non-offensive classification [6]. In this case, a set of non-offensive tweets was a complement of well-defined offensive tweets, i.e., the authors first found offensive tweets by using queries of offensive terms and then built a set of non-offensive tweets by randomly sampling the remaining tweets; the working assumption was that the remaining tweets have a lower offensiveness score. We, on other hand, build rules to identify positive, consolidation speech. In its extreme version, negative sentiment can deteriorate into hate speech which is often directed against identifiable population groups [17, 27] or "trigger events" for hate crime [3]. Sarcasm and irony are often found on Twitter. Their expressions manifest the so-called "bad language" of non-standard spelling, informal vocabulary, slang (*dats why pluto is pluto it can neva b a star*) [8] and negate what is literally said in the text [22]. For the purpose of this study, we consider sarcasm and irony to be nullifiers of positive speech.

Bermingham and Smeaton [2] studied Twitter data during the Irish General Election of 2011. Their annotation categories consisted of three sentiment classes (positive, negative, mixed) and one non-sentiment class (neutral). Tweets deemed un-annotatable, non-relevant, unclear and with contradictory annotations were dismissed from the study. From manually annotated 7 203 documents, only 12% of tweets represented positive sentiments. The authors trained Adaboost MNB classifier and deployed it to classify 32,578 relevant tweets. Identified positive and negative shares of the volume were used as predictors for the election results. Accuracy of prediction varied widely, depending on the time the data was collected. In most cases, the share of positive volume gave more accurate prediction than the negative volume. Interestingly, the manually labeled data had worse predictive power than the classified data. Bollen et al [5] used OpinionFinder and Google Profile of Mood States (GPOMS) to assess predictive power of Twitter data in stock exchange. Publicly available OpinionFinder identifies subjective sentences and positive and negative sentiment expressions, as well as sources of opinion, direct subjective expressions and speech events[4]. GPOMS, built by Bollen et al, measures mood through six dimensions: Calm, Alert, Sure, Vital, Kind and Happy. The dimensions were derived from a rating scale applied to assess multiple dimensions of affect [19]. The authors showed that positivity and Calm better predicted political events and changes in stock market than other variables, although considerable controversy was raised about significance of the results[5]. Empirical evidence shows that messages rebuking hate speech (counter-hate speech) have been posted on Twitter and, albeit very rare, helped to alter aggressive behavior [6]. Although counter-hate speech is promoted as a potent tool of betterment of relations between individuals and groups, there are no reports of studies of counter-hate phenomenon done on massive Twitter data. Thus, our current results can help other researcher in searching for positive speech in large volumes of Twitter data.

---

[2] https://about.twitter.com/company, January 2015

[3] http://orca.cf.ac.uk/68385/

**Semantic Analysis of Positive Speech**

We work with four Twitter data sets collected in Kenya by UMATI project. Each data sets was collected following a violent episode: a grenade attack on a popular Gikomba market (set 1 – 482 tweets), clashes between ethnical/religious groups in Mandera (set 2 – 1156 tweets), assassination of a Muslim cleric Makaburi (set 3– 20,300 tweets), and clashes between ethnical/religious groups in Mpeketoni (set 4 – 106,300 tweets). Our first task was to detect and extract tweets written in English. We used LingPipe's Language Identification[7] and Apache's LangDetect[8] for this task. Our next task was to find unique tweets. (We did not know why those multiple tweets appeared in the data nor we knew representativeness of such repetitions, e.g., were all re-tweets presented or only part of them.) The two operations reduced the data sets as follow: set 1 – 365 tweets, set 2 – 655 tweets, set 3 –8,900 tweets, and set 4 – 41,800 tweets. In semantic analysis of the data, we had to decide on four factors: choice of sentiment categories, sentiment sources in data, sentiment assessment, including interpretation of sarcasm and irony, and sentiment composition.

1. Choice of sentiment categories depends on the goal of sentiment analysis. Sentiments could be polar, e.g., positive and negative in studies of reactions on sport events [10], or multi-categorical, e.g., confusion, gratitude, encouragement in studies of health-related messages [4]. We differentiate between peace-building sentiments (e.g., unity, peace, reconstruction) and support for violence (e.g, anger, intolerance). Peace-building sentiments are the ones that identify positive consolidation speech. Sentiments can be gauged from different parts of tweets: text, URLs, emoticons, hashtags. In some cases tweets are assigned with sentiments of emoticons and hash-tags [16], in others the message content provides the sentiment label [14]. We excluded URLs from the scope of the current study due to uncertainty of their collection. The data did not contain emoticons. Analysis of the data has shown that hashtags were an unreliable source in sentiment identification: in some tweets, hashtags with strong positive sentiments were supported by positive text. In other tweets sentiments were of opposite polarity. Thus, we chose to treat hashtags as regular text.
2. Assessment of sentiments in the message content can be done in several ways. Some apply a composite score of positive and negative sentiments [10], whereas others label text with predominant sentiments [14]. We assess each tweet for presence of positive and negative sentiments.
3. Sentiment composition establishes rules for derivation of sentiments in compound phrases and lexical structures. Some systems consider construction `negative + negative` to be positive, e.g., kill bacteria is labeled positive by TheySay [18]. We, on other hand, mark `negative + negative` as negative.
4. Interpretation of irony and sarcasm depends on topics and subjects of those expressions [22]. We adhere to the notion that sarcasm and irony negate the consolidation aspect of speech; in fact, they can re-make consolidation content into offensive, e.g., adding *HaHa* and *Lol* to R.I.P. (rest in peace) in response to killing of a political figure.

---

[4] http://mpqa.cs.pitt.edu/opinionfinder/
[5] https://folpmers.wordpress.com/2014/04/17/the-twitter-predictor-of-the-dow-the-rise-and-fall/
[6] http://www.ethanzuckerman.com/blog/2014/03/25/susan-benesch-on-dangerous-speech-and counterspeech/
[7] http://alias-i.com/lingpipe/index.html
[8] https://stanbol.apache.org/docs/trunk/components/enhancer/engines/langdetectengine

We hypothesized that examples of positive speech can be very rare, if exist at all. Manual analysis is the most precise tool for extraction and analysis of rare events in text. However, sheer volume of Twitter data prohibits fully manual analysis. We applied semantic semi-manual analysis instead. For each set, we identified content-dependent seed keywords (e.g., *pray, peace*) and used them in bootstrapping to extract more characteristics of positive speech from the remaining data (e.g., *dont celebrate, condolences*). We wanted to evaluate whether WordNet and SentiWordNet can be used to generalize on the found keywords. Unfortunately, the volume of positive speech was too negligent to assess those resources. In terms of speech extraction, we aim to achieve a strong confidence that the identified speech is indeed positive consolidation speech. For this purpose, we seek to populate the target class (i.e., positive speech) with strong examples. Consequently, we accept that the target class misses some of weaker, albeit relevant, examples. In other words, we aim for a higher Precision, even at expense of a lower Recall.

Our semantic analysis was event-based. Each data set discussed one event, and those events begged different reactions from Twitter users. Thus, we studied sentiment categories individually for each set. Strength of the speech and ranking of tweets were outside the scope of this study. Thus, when multiple seed words were present in the same text, this did not affect the resulting label.

Gikomba set covered blasts on Gikomba market, a popular business hub of second-hand clothes. From 365 tweets, most tweets conveyed factual information (e.g., *blasts, victims*); many users showed frustration toward security incompetence and government impotence. We did not find examples of hate or antagonistic speech, only a few tweets allured to a possible terrorist attack. At the same time, not much compassion was expressed either. In this set, we found only three examples of positive speech (<1.0% of the set). Those tweets spoke about reconstruction of the area.

Mandera set was collected in the wake of ethnic and religious clashes in Mandera. In 655 tweets, most tweets reported on inter-clan clashes, and considerable text volume conveyed dissatisfaction with government actions or accused it in escalating the conflict. We found 16 examples of positive speech (< 2.0% of the set). In this case, positive speech called for unity and peace.

Makaburi set consisted of 8,900 tweets commenting on a public assassination of a Muslim cleric Makaburi. Those tweets were strongly separated along ethnic/religious lines. Texts were stronger worded and higher in polarity if compared with the other sets. First, we studied a small sample of data. Majority of tweets belonged to factual type. Aside from them, we found following emotions: considerable anger and appreciation of the killing, including sarcastic tweets; mourning of the killed cleric. Positive speech contained calls for the unity of the nation as a whole, a call for prayer, and a few counter-hate tweets. We selected seed words [*peace, bless, unite, pray, protect, love, support*] and used boostrapping to search the whole set to identify more words and some phrases characteristic of positive speech. After rigorous analysis of extracted tweets, we have been able to identify 59 examples of positive speech (< 0.7% of the set).

Mpeketoni set, with 41,800 tweets, has been the largest data by far. Tweets were about a series of Mpeketoni attacks, blamed on Al Shabaab (Islamist militia) and the area's proximity to Somali. Aside of neutral and factual tweets, we identified the following categories of tweets: frustration with cycles of violence; blame of government, including sarcastic tweets; accusation and anger. Positive speech had calls for unity, calls for law and order, calls for prayer, and a few reconstruction and counter-hate tweets. We identified positive seed words [*peace, pray, unite, condolence, god, condemn, thoughts, reclaim, one* ] and used them to find other words and phrases which appear in possibly positive tweets. After 2nd round of extracting tweets and their manual analysis, we found 320 examples of positive speech ( < 0.8% of the sets).

**Topic Modeling by Latent Dirichlet Allocation**

Positive consolidation speech is shown to be most effective during a short window frame or operating in certain localities [15,28]. Identifying it within a short time-frame in a noisy Twitter environment can be a daunting task. A few topic-based models have exhibited advanced results. In these approaches a group of topics, is learned from the raw (ex. Bag-of-words) representation of the text. The document is then represented by the projection into the topic space. These approaches include Latent Semantic Indexing (LSI) and its probabilistic version, probabilistic Latent Semantic Indexing. A more comprehensive approach is based on the Latent Dirichlet Allocation (LDA) model [12].

LDA is an unsupervised Machine Learning technique that finds latent topics in large collection of documents. Each document is represented as a probability distribution $\theta$ over some topics, while each topic is represented as a probability distribution $\varphi$ over a number of words; $\theta$ and $\varphi$ are assumed to have Dirichlet prior. For each document, LDA picks a topic from $\theta$, samples a word from the distribution over the words associated with the chosen topic, repeats the process for all the words in the document [12]. The method has been successfully used in topic identification in Twitter, including sentiment analysis [9].

To output meaningful results, the method requires considerable tuning, e.g., selection of the number of topics, data pre-processing. We tested removal of stop-words, but decided against it after topic quality remarkably decreased. For each data set, we worked with 5,10,15,20, and 25 topics. For the largest Mpeketoni set, we additionally worked with 35, 45 and 50 topics. From Text Data Mining perspective, LDA results were reliable and representative of the data sets contents. For all the data sets, LDA built several coherent topics that answered "what", "where" and/or "who" questions related to the main content of the set.

On each set, LDA required different # of topics to produce the most logical and articulated topics: Gikomba set: the most descriptive and informative topics were built with N= 10, 15. Individual topics became less informative, e.g., omitted location or the event, and more fragmented for N = 20, 25. Example of a topic: *the, of, Gikomba, market, http:, Nairobi, by, at, blasts, fire, has, victims, area, RT, The, for*

Mandera set: LDA with N = 10, 15 produced better results. Again, individual topics became less informative with N = 20,25.  Example of a topic*: in, #Mandera, RT, as, #Wajir, militia, clan, are, not, that, tribal, why, clashes, there, people , their, armed, war, were*

Makaburi set: N = 10, 15, 20 helped to build most descriptive and interesting topics. LDA performance declined for N = 25.  Example of a topic*: Makaburi, shot, in, dead, cleric, Abubakar, Muslim, Sheikh, Mombasa, RT, has, alias, Shariff, Radical, been, #Makaburi, radical, is, dead., muslim*

Mpeketoni set: it was the largest set. We obtained coherent topics with N = 35, 45, 50. For smaller N = 5, ..., 25, the resulting topics omitted part of essential information.   Example of a topic: *in, Mpeketoni, attack, Kenya, people, 48, of, town, killed, on, #Kenya, Kenyan, least, at, #Mpeketoni, near, coastal, dead, 50*

LDA also annotated every tweet with words that can be found in topics and the corresponding topic number. That made analysis of the data feasible. For example, we were interested in tweets which say about victims of the Gikomba explosions. After a straight-forward search, we found *victims*(6) in tweets # 12, 97, 137, etc. A critical advantage of LDA was its ability to find topics supported by a small portion of text and in considerable semantic distance from the mainstream topics (i.e., topics- outliers).

In data sets Mandera and Mpeketoni, LDA was able to identify topics of peace (*and, wajir, mandera, of, should, be, peace, both, security, we, leaders, solution, must, @jageyo, communities, that, who, #KTNBottomline, well, In*) and condolences (*the, to, of, in, and, those, My, who, their, lost, families, condolences, all, #MpeketoniAttack, God, my, for, victims, affected, May*) despite small volumes of those texts. In Makaburi set, LDA was able to identify sarcasm as a separate topic (*RT, Makaburi, MAKABURI, R.I.P., #MAKABURI, RIP, A, Man, The, Al, (cont), Lol, ?, Good, IS, TO*). This is remarkable achievement, given that sarcastic tweets appeared extremely rare in the data. LDA, however, was able to pick up self-contradictory combinations of *R.I.P, lol, good* in those tweets.  After analysis of the built topics, we again were able to identify tweets that deliver these messages. For example, in Mandera set, *peace*(7) appeared in tweets # 20, 35, 49, 210, etc.

 The output topics summarized main content in four separate Twitter data. On three data sets, the output included topics presented by a small number of text (<0.7%). Note that LDA obtained those results on data that exhibited characteristics of online " bad language": slang, spelling and grammatical variations, short-comings [8]. Critically, LDA's analysis of tweets supports and independently validates the results obtained by semantic analysis.

**Text Representation: Word Vectors vs N-Grams**

Bag-of-words (BOW) is a basic, classic model for text representation. In it, a document is represented by a vector of its words' frequencies normalized over a set of documents. Despite of its simplicity this representation is still widely used for a plethora of different tasks (classification, information retrieval and others) and has been proved very successful. The two biggest drawbacks of this model are its extreme sparseness, particularly for short text such as tweets, and its lack of a representation for relationship between words. An N-gram model, is a natural extension of the BOW. It is based on the fact that the presence of a sequence of N words, an N-gram, does not necessarily carry the same semantic weight as the presence of its constituents. This model represents a document not only by the frequency of its words, but also by the frequencies of its N-grams, up to a certain value of n. In practice values of n=2, or n=3 are used. These models have been used for many years, and have shown great results, despite the fact that they suffer ever more from the extreme sparseness problem of the BOW model. In our study, we compare a) BOW or 1-gram, represents the document by the normalized frequencies of the document words, with stop words removed; b) 2,3-grams models represent the document by the normalized frequencies of the document words and 2-word phrases (or 2-word and 3-word phrases) which are formed without removing the stop words; c) NB-SVM is the exact algorithm presented in [25] which by combining the generative and discriminative classifiers provides, in resume, an Support Vector Machine.

Recently there have been developments in the distributed word representation. These representations are learned from a text corpus by a neural network model whose goal is to predict the next word given a certain context. Words, similar both grammatically and semantically, tend to appear in similar context and thus get assigned similar co- appearance coefficients. The resulting word vectors can be considered as a mapping of a word into an N-dimensional space, where linguistic regularities can be observed. Such a representation is a perfect illustration of a famous quote "You shall know a word by the company it keeps (Firth, J. R. 1957:11)". Since their introduction, these models have been applied to a multitude of applications, including text classification [29,30].

In classification experiments, we used original data sets, without extracting English tweets first. We chose to preserve as much of the original information in the tweet as possible and kept the pre-processing to a minimum: the links replaced by a tag URL, all usernames were replaced by a tag USER and all "#" were stripped from the hashtags. Punctuation was separated from words and kept. Finally, the text was converted to lower case. We used an off-shell python implementation of the skip-gram model provided by the gensim package [9]. Training parameters were crucial in the applicability of the resulting model. These were found manually using words association (see next section) as feed-back. For both types of datasets, the best parameters were: no negative sampling, a sampling threshold of 0.001 and a min count of 1.

Distributed word models have been so far obtained on rather large datasets. In this work, we are dealing with datasets that are as small as 1000 samples (Gikomba set), so the pertinence of the built model has to be verified prior to its use in a data mining task. We performed this verification manually, using the fact that the model places semantically similar words close in the embedded space, allowing us to probe the model locally by constructing word associations for a few query words and to check that the surrounding embeddings are meaningful. If successful, we probe the model globally by running a clustering algorithm on the word embeddings and verifying the relevancy of the resulting clusters. Clusters from Mpeketoni sets include: *hanging handling wednesday ruthless golden hurt 98 pursue masterminds*, *educate whaaaat cow veteran resettle mau landless 1960 settlemen*[10]

---

[9] http://is.muni.cz/publication/884893/en
[10] Reference to re-settlement of Kenyan tribes in 1960-ies.

We query the model with a word or an ensemble of words that would represent a topic in the dataset and check the near-by embeddings, by looking at the most similar words in the model according to cosine similarity. For example for Mpeketoni sets - a dataset dealing with a bombing likely by a terrorist organization Al Shabaab - a query on "Al-Shabaab" would be meaningful. The 10 most similar words to it were: *claim, attacks, fake, shabab, discussion, twitter, responsibility, rejects, naïve.* As we see, the response contains words relating to attack, responsibility, and the name of the group. These words are relevant and we can conclude that the model did capture the semantics of the corpus.

To further test the quality of the model we performed Dirichlet Process Gaussian Mixture Model (DPGMM) based clustering on the embedded vocabulary, with the expectation that semantically or grammatically similar words would form some meaningful clusters. Gaussian Mixture Models (GMM) have already been used in text classification based on distributed vector models with very good results [29]. We used the DPGMM model from the scikit-learn toolkit[11] with the diagonal covariance and a maximum of thirty clusters. To speed up the process, we only used 9000 medium-frequency words from the total corpus vocabulary.

The word vector model provides us with a way of representing individual words, but these individual vectors need to be combined to produce a text vector of a constant length for the entire document. In other words, a proper pooling method of representing a multi-set of word vectors into a single high dimensional vector needs to be found. Common techniques concentrate on both representation of the centroid (i.e., mean-vector pooling) and on distribution of word vectors. This has been achieved by means of a class specific or global Gaussian Mixture Models (GMM)[29]. Our approach differs from the above in that we do not fit a distribution model to the multi-set of word vectors, but rather employ statistical analysis measures such as averaging and standard deviation to describe this ensemble and its transformations.

Let the matrix M denote a multi-set of word vectors, where the rows, $M_i$, represent the words of the tweet in the order they appear and columns the dimensions in the embedded space. For a tweet of length N and a d-dimensional model, M is a N× d matrix. The rows in this matrix define an ordered set of points, a path, in the d-dimensional space. For this multi-set we can easily define a mean µ and a standard deviation σ as a standard deviation of the ensemble of vectors, $M_i$ (rows), constituting the set M:

$$\mu = \frac{1}{N}\sum_{i=1}^{N} M_i, \quad \sigma = \sqrt{\frac{1}{N}\sum_{i=1}^{N}(M_i - \mu)^2}$$

These two features µ, σ describe the centroid and the spread of the set of points in the d-dimensional space defined by the word vectors.

**Data Classification**

Our classification experiments consisted of 5 repetitions. In each repetition a word vector model was trained on all of the data. The full dataset would then be vectored using some of the features described in above. The resulting matrix was split into a training and testing sets according to a 5-fold cross validation and passed to a linear regression classifier (LR) and a support vector machine (SVM). The resulting accuracy, precision, recall and the f-score for the minority class were all recorded for each of the 25 total runs. The experiment was repeated for different sets of features and different datasets. We used the classification framework from scikit-learn toolkit[11].

---

[11] http://scikit-learn.org/stable/

Due to the nature of the Kenyan datasets, positive speech is extremely rare (as low as a few percent of the data) and thus finding such tweets presents a challenge as much for a human as for a classifier. Before venturing into a full study, we looked closely at the results of the classification of our largest in- house labelled Mpeketoni set, using µ,σ features. The results were: 538 true positives, and 301 false positives. Upon manual inspection of the 301 false positives we discovered that actually 210 of them were indeed real positives. We then implemented an iterative procedure, in which we inspected and relabelled the incorrect false positives found by the classification algorithm. With this simple procedure, in only four iterations we increased the number of positively labelled instances by almost 50%, 965 positive tweets in total. Some of the missed tweets were the incorrectly labelled re-tweets of correctly labelled messages. However there were equally many new original examples of positives phrases missed by the human and identified by the algorithm:

*In the end we all Africans..let the Games Roarr.#TeamAfrica,*
*My God of peace and mercy comfort the people of mpeketoni*

On the next step, we launched an investigation of the effect of different features on classification results. We pay a special attention to training time of the word-to-vector models, as this can critically influence their applicability - see Table 1. We repost classification F-score in Table 2; Table 3 reports on time spent on classification. Empirical evidence showed that SVM consistently obtained a better accuracy on N-gram based features than on word-embedding features; LR obtained better accuracy on distributed word representations. As we expected, time efficiency favored 1-gram text representation.

**Table 1: Model training time (in secs).**

|  | Mandera | Makaburi | Mpeketoni |
|---|---|---|---|
| # of tweets | 1,200 | 30,000 | 106,000 |
| w2v training time |  | 72 | 1680 |

**Table 2: F-score obtained on different feature sets. Best results are in bold; best ties are in bold italics.**

|  | Linear Regression | | | Support Vector Machine | | |
|---|---|---|---|---|---|---|
| Features | Mandera | Makaburi | Mpeketoni | Mandera | Makaburi | Mpeketoni |
| µ | 0.57 | **0.56** | 0.50 | 0.48 | 0.54 | 0.48 |
| µ, σ | **0.62** | 0.55 | *0.62* | 0.55 | 0.47 | 0.63 |
| NB-SVM | 0.44 | 0.43 | 0.54 | 0.43 | 0.44 | 0.54 |
| 1-Grams | 0.31 | 0.26 | *0.62* | *0.64* | 0.55 | 0.72 |
| 1,2-Grams | 0.45 | 0.21 | *0.62* | *0.64* | 0.56 | *0.76* |
| 1,2,3-Grams | 0.45 | 0.21 | 0.61 | *0.64* | **0.57** | *0.76* |

**Table 3: Classification time of different feature sets (in secs). Best results are in italic.**

|  | Linear Regression | | | Support Vector Machine | | |
|---|---|---|---|---|---|---|
| Features | Mandera | Makaburi | Mpeketoni | Mandera | Makaburi | Mpeketoni |
| µ | 0.09 | 2.88 | 12.60 | 0.06 | 7.11 | 1458.00 |
| µ, σ | 0.21 | 5.98 | 33.60 | 0.09 | 6.38 | 3261.00 |
| NB-SVM | 0.15 | 6.93 | 81.73 | 0.29 | 17.50 | 1474.00 |
| 1-Grams | *0.02* | *0.25* | *2.15* | *0.08* | *3.63* | *182.50* |
| 1,2-Grams | 0.03 | 0.49 | 5.22 | 0.13 | 7.27 | 528.60 |
| 1,2,3-Grams | 0.03 | 0.84 | 9.60 | 0.15 | 9.42 | 795.70 |

**Classification of Publicly Available Data**

We replicated our experiments on publicly available Twitter data. The dataset[12] contains 1.6 million emoticons-containing tweets on a variety of subjects from gadgets, products, people and events. The tweets were labelled according to positive, negative or neutral sentiment conveyed by the emoticons; emoticons were later stripped off the tweet. A small, hand-labelled test set was provided. We randomly extract subsets of 0.1% (1,600 tweets), 1% (16,000) and 10% (160,000) from the full dataset and run the classification experiment in the same manner as described above. The random sampling of tweets for labelled dataset was done with care to preserve the original balance of the data (50 / 50). The qualitative dependence of the f-score's behaviour for various feature set for different training corpus sizes is similar to that found in Kenyan data. For 0.1% of tweets the mean-vector is the best, similar to Mandera. For 1% dataset, roughly of the same size as the Makaburi dataset, all features in the table are informative, but the f-score increase is small. For 10% of the data - the same size as the Mpeketoni dataset, the effect of complex features increasing the classification is clearly visible, as it is for the full dataset. Tables 4- 6 report the results.

The goal of this study was to investigate the use of novel text representation techniques in classification of Twitter data sets; the data is challenging for automated analysis. As such we did not make any effort in optimizing the classification results per se, and reported the results for only one end-classifier (LR). We obtain very similar dependencies on all mode parameters with other base classifiers (SVM)[13], however the actual numbers for f-score differ. Nonetheless in order to compare our results with the previous work on the sentiment dataset we performed experiments in the similar manner: we trained our model on the full dataset, and the tested it on the test set provided. We observed accuracies ranging from 81% to 83% consistent with that previously reported.

**Table 4: Model training time (in secs).**

|  | 0.1% tweets | 1% tweets | 10% tweets |
|---|---|---|---|
| # of tweets | 1,600 | 16,000 | 160,000 |
| w2v training time | 5 | 180 | 2100 |

**Table 5: F-score obtained on different feature sets. Best results are in bold; best ties are in bold italics.**

|  | Linear Regression | | | Support Vector Machine | |
|---|---|---|---|---|---|
| Features | 0.1% tweets | 1% tweets | 10% tweets | 0.1% tweets | 1% tweets |
| $\mu$ | 0.64 | 0.69 | 0.72 | 0.64 | 0.69 |
| $\mu, \sigma$ | 0.64 | 0.71 | 0.74 | 0.64 | 0.70 |
| NB-SVM | 0.55 | 0.74 | 0.74 | 0.48 | 0.72 |
| 1-Grams | *0.69* | *0.76* | 0.79 | *0.68* | *0.75* |
| 1,2-Grams | *0.69* | *0.76* | *0.80* | *0.68* | **0.76** |
| 1,2,3-Grams | *0.69* | *0.76* | *0.80* | *0.68* | 0.75 |

---

[12] http://twittersentiment.appspot.com, Alec Go, Richa Bhayani, and Lei Huang. 2009. Twitter sentiment classification using distant supervision

[13] We were able to run SVM on 0.1% Twitter and 1% Twitter sets, but not on the 10% Twitter set.

Table 6: Classification time of different feature sets (in secs). Best results are in italic.

|  | Linear Regression | | | Support Vector Machine | |
| --- | --- | --- | --- | --- | --- |
| Features | 0.1% tweets | 1% tweets | 10% tweets | 0.1% tweets | 1% tweets |
| μ | 0.09 | 1.22 | 11.40 | 6.82 | 341.00 |
| μ, σ | 0.31 | 2.39 | 22.50 | 17.35 | 1527.00 |
| NB-SVM | 0.25 | 7.29 | 179.00 | 3.72 | 903.00 |
| 1-Grams | *0.03* | *0.26* | *5.51* | *0.78* | *75.00* |
| 1,2-Grams | 0.04 | 0.44 | 9.60 | 0.95 | 102.00 |
| 1,2,3-Grams | 0.05 | 0.60 | 11.50 | 0.95 | 109.00 |

**Discussion and Future Work**

In this study, we presented a method for analysis of positive speech on Twitter. Albeit occurring on Twitter, positive speech is severely under-represented when the data is collected from conflict zones (e.g., < 1% of collected data). We have shown that despite the severe imbalance, automated methods are capable of identification and classification of positive speech. We have shown that LDA can successfully identify main topics in each data (i.e., factual information), as well as find and extract topics supported by a small number of tweets (i.e., positive speech, sarcasm). We have compared N-gram based text representation with a novel technique of representing tweets as a vector based on distributed word models. In our experiments, N-grams outperformed distributed word models. To see the benefit from the distributed word models, datasets on the order of $10^5$ samples are required. There is also interplay between the size of labeled data and dimensionality of the word representation; increasing one or the other leads to a better classification results, however, increasing both is not necessary. In the future we would like to apply this technique for severely imbalanced data classification and rare event detection in short texts, as well as use this representation for texts visualization.

In study of positive speech on Twitter, we see following directions of future work: a) test other methods of topic identification in Twitter data, for example, non-negative matrix factorization which is able to find topics-outliers in Twitter data [23]; b) use multiple-view learning [11] to analyze positive and negative speech in Twitter data; c) apply fine-grained semantic analysis of counter- hate speech.

**Acknowledgements**

We thank UMATI project   providing the data sets. This study was in part supported through NSERC Discovery grant available, NSERC CREATE grant and Polish National Scientific Centre NCN grant.

**References**

[1] P. Barbera and G. Rivero. Understanding the political representativeness of twitter users. Social Science Computer Review, to appear.
[2] A. Bermingham and A. Smeaton. On using twitter to monitor political sentiment and predict election results. In Proceedings of SAAIP 2011, pages 2–10,2011.
[3] M. L. Best. Peacebuilding in a networked world. Commun. ACM, 56(4):30–32, 2013.
[4] V. Bobicev, M. Sokolova, and M. Oakes. Recognition of sentiment sequences in online discussions. In Proceedings of SocialNLP 2014, pages 44 – 49, 2014.
[5] J. Bollen, H. Mao, and X. Zeng. Twitter mood predicts the stock market. Journal of Computational

Science , 2(1):1–8, 2011.

[6] W. Campbell, E. Baseman, and K. Greenfield. Content+context=classification: Examining the roles of social interactions and linguist content in twitter user classification. In Proceedings of SocialNLP 2014 ,pages 59 – 65, 2014.

[7]L. Conway and M. Schaller. How communication shapes culture. In K. Fiedler, editor, Social Communication , pages 107 – 127. New York: Psychology Press, 2007.

[8] J. Eisenstein. What to do about bad language on the internet. In Proceedings of NAACL-HLT, pages 359 – 369, 2013.

[9] M. Erdmann, K. Ikeda, H. Ishizaki, G. Hattori, and Y. Takishima. Feature based sentiment analysis of tweets in multiple languages. In Web Information Systems Engineering–WISE 2014 , pages 109–124. Springer, 2014.

[10] A. Gruzd, S. Doiron, and P. Mai. Is happiness contagious online? a case of twitter and the 2010 winter olympic. In System Sciences (HICSS) , pages 1– 9. IEEE, 2011.

[11] X. He, M.-Y. Kan, P. Xie, and X. Chen. Comment-based multi-view clustering of web 2.0 items. In Proceedings of the 23rd WWW Conference , pages 771–782, 2014.

[12] L. Hong and B. D. Davison. Empirical study of topic modeling in twitter. In Proceedings of the First Workshop on Social Media Analytics, pages 80–88. ACM, 2010.

[13] N. A. Kimbrel, R. O. Nelson-Gray, and J. T. Mitchell. Bis, bas, and bias: The role of personality and cognitive bias in social anxiety. Personality and Individual Differences , 52:395–400, 2014.

[14] S. Kiritchenko, X. Zhu, and S. M. Mohammad. Sentiment analysis of short informal texts. Journal of Artificial Intelligence Research , 50:723–762, 2014.

[15] W. A. Knight. Evaluating recent trends in peacebuilding research. International Relations of the Asia-Pacific , 3(2):241–264, 2003.

[16] E. Kouloumpis, T. Wilson, and J. Moore. Twitter sentiment analysis: The good the bad and the omg! In Proceedings of the Fifth International AAAI Conference on Weblogs and Social Media, pages 538–541. AAAI, 2011.

[17] I. Kwok and Y. Wang. Locate the hate: Detecting tweets against blacks. In Proceedings of AAAI , pages 1621–1622. AAAI, 2013.

[18] K. Moilanen and S. Pulman. Sentiment composition. In Proceedings of Recent Advances in Natural Language Processing , pages 378–382, 2007.

[19] J. Norcross, E. Guadagnoli, and J. Prochaska. Factor structure of the profile of mood states (poms): two partial replications. Journal of Clinical Psychology,40(5):1270–1277, 1984.

[20] G. Ou, W. Chen, T. Wang, Z. Wei, B. Li, D. Yang, and K.-F. Wong. Exploiting community emotion for microblog event detection. In Proceedings of EMNLP 2014, pages 1159–1168, 2014.

[21] A. Razavi, D. Inkpen, S. Urisky, and S. Matwin. Offensive language detection using multi-level classification. In Advances in Artificial Intelligence, pages 16–27. Springer, 2010.

[22] P. Rosso. Socialirony. In Proceedings of SocialNLP 2014 , page Invited talk, 2014.

[23] G. Shen, W. Yang, W. Wang, M. Yu, and G. Dong. Detecting anomalies in microblogging via nonnegative matrix tri-factorization. In Social Media Processing, pages 55–66. Springer, 2014.

[24] Z. Tufekci. Big questions for social media big data: Representativeness, validity and other methodological pitfalls. In ICWSM, 2014.

[25]S. Wang and C. Manning. 2012. Baselines and bigrams: Simple, good sentiment and topic classification. In Proceedings of the 50th Annual Meeting of the Association for Computational Linguistics: Short Papers - Volume 2, ACL '12, pages 90–94, Stroudsburg, PA, USA. Association for Computational Linguistics.

[26] W. Warner and J. Hirschberg. Detecting hate speech on the world wide web. In Proc. of the 2012 Workshop on LSM, pages 19 – 26, 2012.

[27] P. Watzlawick, J. B. Bavelas, and D. D. Jackson. Pragmatics of human communication: A study of interactional patterns, pathologies and paradoxes. WW Norton and Company, 2011.

[28] H. won Jeong. Peacebuilding in Postconflict Societies: Strategy and Process. Lynne Rienner Publishers, London, 2005

[29] C. Xing, D. Wang, X. Zhang, and C. Liu. 2014. Document classification with distributions of word vectors. In Asia-Pacific Signal and Information Processing Association, 2014 Annual Summit and Conference (APSIPA), pages 1–5. IEEE.

[30] M. Yu and M. Dredze. 2015. Learning composition models for phrase embeddings. Transactions of the Association for Computational Linguistics, 3:227–242